\documentclass[conference]{IEEEtran}
\IEEEoverridecommandlockouts

\usepackage{cite}
\usepackage{amsmath,amssymb,amsfonts}
\usepackage{graphicx}
\usepackage{textcomp}
\usepackage{xcolor}
\usepackage{hyperref}
\usepackage{algorithm,algpseudocode}
\usepackage{amsmath}
\usepackage{subcaption,multirow}
\def\BibTeX{{\rm B\kern-.05em{\sc i\kern-.025em b}\kern-.08em
    T\kern-.1667em\lower.7ex\hbox{E}\kern-.125emX}}
\begin{document}

\title{Adaptive Aspect Ratios with Patch-Mixup-ViT-based Vehicle ReID}

\author{\IEEEauthorblockN{%
  Mei Qiu,  Lauren Ann Christopher, Stanley Chien, Lingxi Li$^*$\thanks{$^*$Corresponding Author}}%
\textit{Purdue University, USA.} \\ 
\{qiu172, lachrist, yschien, lingxili\}@purdue.edu}

\maketitle

\begin{abstract}
Vision Transformers (ViTs) have shown exceptional performance in vehicle re-identification (ReID) tasks. However, non-square aspect ratios of image or video inputs can negatively impact re-identification accuracy. To address this challenge, we propose a novel, human perception driven, and \textbf{general} ViT-based ReID framework that fuses models trained on various aspect ratios. Our key contributions are threefold: (i) We analyze the impact of aspect ratios on performance using the VeRi-776 and VehicleID datasets, providing guidance for input settings based on the distribution of original image aspect ratios. (ii) We introduce patch-wise mixup strategy during ViT patchification (guided by spatial attention scores) and implement uneven stride for better alignment with object aspect ratios. (iii) We propose a dynamic feature fusion ReID network to enhance model robustness. Our method outperforms state-of-the-art transformer-based approaches on both datasets, with only a minimal increase in inference time per image.

The code is released here: \href{https://github.com/qiumei1101/Adaptive_AR_PM_TransReID.git}{\color{blue}Adaptive\_AR\_PM\_TransReID}.
\end{abstract}


\section{Introduction}
\label{sec:intro}
Vehicle re-identification (ReID) is critical in intelligent transportation systems, tasked with identifying vehicles across multiple non-overlapping cameras \cite{wang2019survey}. Vehicle Re-ID has seen success with both Convolutional Neural Network (CNN) \cite{he2016deep, bashir2019vr, roman2021improving} and Vision Transformers (ViTs) backbones \cite{dosovitskiy2010image, he2021transreid, lian2022transformer, luo2021empirical, wei2022transformer}. 
Challenges such as variations in viewpoint, pose, illumination, occlusion, background clutter, similar vehicle models, resolution differences, temporal changes, aspect ratios, and privacy constraints, all contribute to the complexity of obtaining accurate identification. Addressing these issues requires deep learning models to extract robust and discriminative features that can withstand these variations \cite{zheng2020vehiclenet, chen2023global}. A combination of global and local features is essential for effective vehicle representation \cite{peng2019learning, zhang2020part, gu2021efficient}. While distinct identifiers such as license plates are accurate and widely used in urban areas \cite{ramajo2024dual, amiri2024comprehensive}, they are less applicable in the highway context due to privacy concerns and practical limitations. 

Numerous benchmark datasets, such as VeRi-776 \cite{zheng2020vehiclenet}, PKU-VD \cite{yan2017exploiting}, VehicleID \cite{liu2016deep}, 
and VERI-Wild \cite{lou2019veri}, are crucial for developing algorithms for vehicle ReID  \cite{zakria2021trends}. State-of-the-art models leverage self-attention mechanisms, with ViTs  capturing discriminative details better than CNN-based methods \cite{he2021transreid}. ViTs adapt transformers from NLP \cite{vaswani2017attention} to computer vision, using self-attention on image patches for tasks like image classification. Variants such as DeiT \cite{touvron2021training}, Swin \cite{liu2021swin}, and PVT \cite{wang2021pyramid} have proven effective in image classification, detection, and Re-ID \cite{khan2022transformers}. However, the varying aspect ratios across datasets present challenges in model training.
Unlike CNN-based models, ViTs treat the entire image as a sequence of patches, necessitating careful resizing and cropping \cite{ke2021musiq, liu2022aspect, mao2022towards, dehghani2023patch}. Early ViT implementations adopted resizing strategies from CNNs, often distorting aspect ratios and potentially compromising performance, especially in tasks dependent on object shape and scale.
Subsequent studies explored padding strategies, adaptive post-patch extraction, trainable resizing networks, aspect-ratio aware attention mechanisms, and multi-scale/multi-aspect training \cite{xia2022vision, lv2022scvit, hwang2022vision, zhu2022aret, ke2021musiq, li2022multi}. However, these approaches introduce computational burdens, data requirements, and optimization challenges.

\begin{figure}
    \centering
    \includegraphics[width=0.48\textwidth]{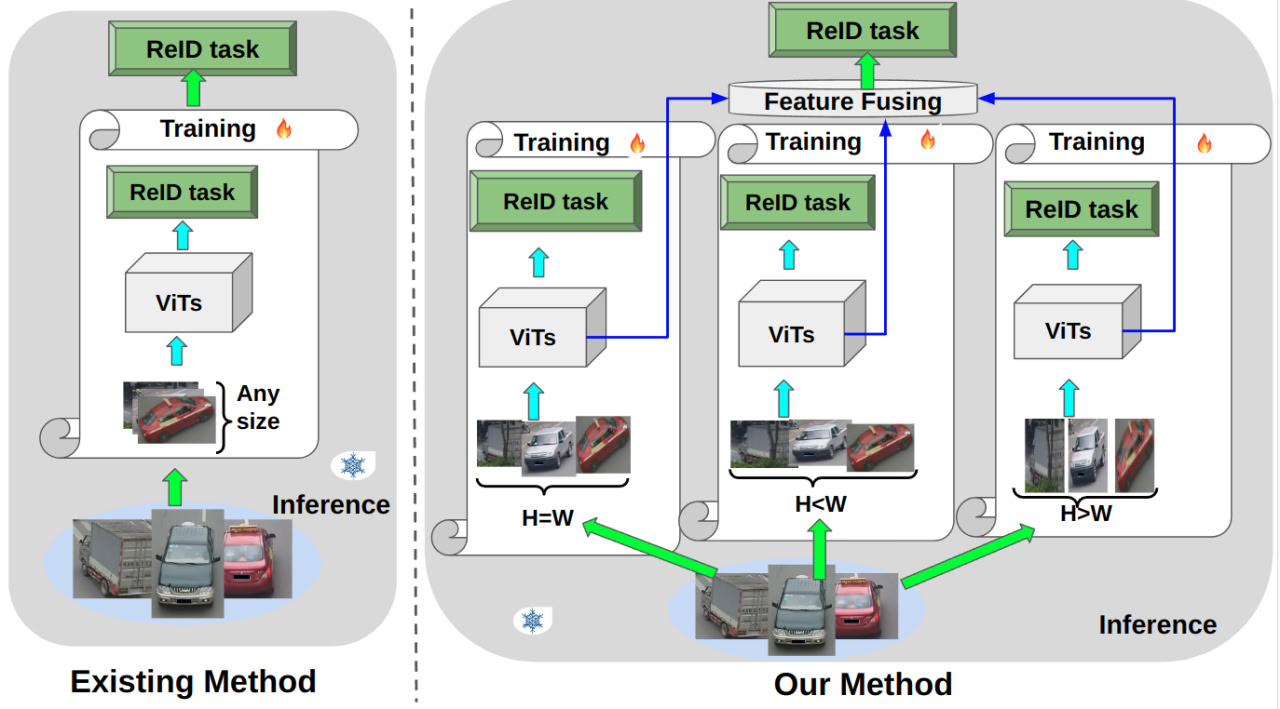} 
    \vspace{2mm}
    \caption{\textit{ (Left) Existing Method: Image size is typically fixed and set to a single square shape. (Right) Our Method: Combined Vision Transformer (ViT)-based ReID model that dynamically fuses features extracted from multiple models. Each model is trained on a different fixed size and aspect ratio.}}
    \label{fig:contrib}
    \vspace{2mm}
\end{figure}

In summary, there is a research gap in applying ViTs to vehicle ReID, including optimizing scaling and resizing strategies, understanding aspect ratio effects, and exploring patch-level mixup \cite{zhang2018mixup} in multi-aspect ratio scenarios. Intra-image mixup can help models learn detailed features despite resizing distortions. For instance, TransReID \cite{he2021transreid} introduced a jigsaw patch module (JPM) to enhance feature robustness, but pixel-level intra-mixup remains unexplored.

\begin{figure}
    \centering
    \includegraphics[width=0.48\textwidth]{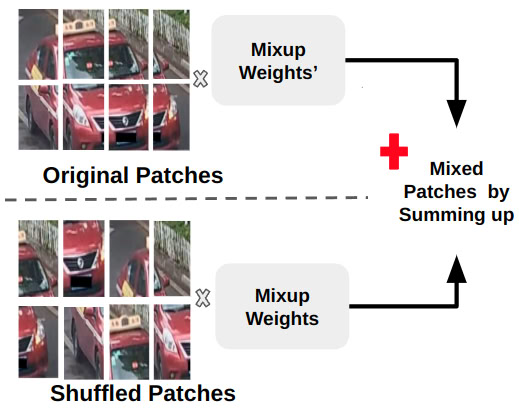} 
    \vspace{2mm}
    \caption{\textit{PM module. }}
    \label{fig:pm}
    \vspace{2mm}
\end{figure}

We propose aspect ratio as a key factor affecting vehicle ReID performance and the robustness of feature learning in ViTs. We conduct comprehensive experiments to explore the effects of various aspect ratios on ViT-based ReID.

To enhance the model's generality to various aspect ratios, we dynamically fuse features from several models trained on images with different aspect ratios, as shown in Fig. \ref{fig:contrib}. Additionally, we propose a novel intra-image patch mixup (PM) data augmentation method to improve the model's learning of details and mitigate overfitting during training. We also employ an uneven stride strategy in the patchify step to reduce distortion caused by resizing.

The key contributions of this work are summarized as follows:
\begin{itemize}
    \item Identified aspect ratio as a critical factor affecting vehicle ReID performance and ViT's feature learning robustness.
    \item Conducted extensive experiments on the impact of aspect ratios on ViT-based ReID.
    \item Developed a general ViT-based ReID framework that uses dynamic feature fusion across different aspect ratios to enhance robustness.
    \item Introduced intra-image patch mixup (PM) to improve learning and reduce overfitting.
    \item Implemented an uneven stride strategy to reduce distortion from resizing.
\end{itemize}


\section{Proposed Method}
\label{sec:pagestyle}
\subsection{Model Structure}
We train separate models for each major aspect ratio ($H=W, H<W, H>W$) as shown in Fig. \ref{fig:contrib}. Resized inputs for each model are augmented with Patch Mixup (PM) (Fig. \ref{fig:pm}). A pre-trained Vision Transformer (ViT) backbone initializes the models, and features from the last transformer layer are used for the ReID task.

\smallskip
\noindent
\textbf{Training Pipeline}

We optimize the network by constructing ID loss and triplet loss for global features from the class token in ViTs with the same weight of $1.0$. The ID loss $\mathcal{L}_{\text{ID}}$ is the cross-entropy loss without label smoothing. For a triplet set $\{a, p, n\}$, the triplet loss $\mathcal{L}_{\text{tri}}(\theta; X)$ with a soft-margin can be calculated as:
\begin{equation}
\mathcal{L}_{\text{ID}}(\theta; x) = -\frac{1}{P \times K} \sum_{i=1}^{P \times K} \log \left( p\left( y_i \mid x_i \right) \right),
\end{equation}

\noindent where $P \times K$ is the batch size, $x_i$ refers to the image sample of vehicle $i$, and $y_i$ denotes its ID label.

\begin{align}
\mathcal{L}_{\text{tri}}(\theta; X) = \frac{1}{P \times K} 
& \sum_{i=1}^{P} \sum_{a=1}^{K} \log \left[ 1 + \exp \left( \vphantom{\min_{\substack{j=1 \dots P, n=1 \dots K, j \neq i}}}\right.\right. \notag \\
& \quad \left.\left.\max_{p=1 \dots K} D\left( f_\theta\left( x_a^i \right), f_\theta\left( x_p^i \right) \right) \right. \right. \notag \\
& \quad \left. \left. - \min_{\substack{j=1 \dots P, \\ n=1 \dots K, \\ j \neq i}} D\left( f_\theta\left( x_a^i \right), f_\theta\left( x_n^j \right) \right) \right) \right]
\end{align}

\noindent where $f_\theta\left( x_a^i \right)$ refers to the anchor sample, $f_\theta\left( x_p^i \right)$ indicates the hard positive sample, and $f_\theta\left( x_n^j \right)$ denotes the hard negative sample. $D$ denotes the squared Euclidean distance. The overall loss is:
\begin{equation}
L_{\text{over}}(\theta; x) =L_{\text{ID}}(\theta; x) + L_{\text{tri}}(\theta; X).
\end{equation}

\smallskip
\noindent
\textbf{Inference Pipeline}

We use a \textbf{Dynamic Feature Fusing} strategy during inference to fuse features from models trained on multiple inputs. The output feature vector can be represented as:
\[
\textbf{F\_out} =  w_1 \cdot \textbf{f}_1 + w_2 \cdot \textbf{f}_2 +  w_3 \cdot \textbf{f}_3,
\]

\noindent where \(w_1\), \(w_2\), and \(w_3\) are weights. For the adaptive weight assignment \(w\):

\[
w = 
\begin{cases}
    1.3, & \text{if } |\text{model\_ar} - \text{image\_ar}| \leq 0.3 \\
    1.0, & \text{if } 0.3 < |\text{model\_ar} - \text{image\_ar}| \leq 0.6 \\
    0.9, & \text{otherwise}
\end{cases}
\]

\noindent Here, \textit{model\_ar} is the aspect ratio (AR) of the input image size the model was trained on, and \textit{image\_ar} is the AR of the image during inference. The AR is the width-to-height ratio. The difference between \textit{model\_ar} and \textit{image\_ar} is used to assign weights \(w\), adjusting feature influence from different models to enhance ReID robustness. For different datasets, the best $w$ can be slightly different.

\subsection{Key Features}
\smallskip
\noindent
\textbf{Image Input with Adaptive Size and Shape.} We statistically estimate the dataset's size and aspect ratios. Using the mean or median size is effective, and K-means clustering can identify key aspect ratio clusters. This approach informs resizing images to train each model on a fixed size and aspect ratio.

\smallskip
\noindent
\textbf{Patchification with Uneven Stride.} To enhance spatial learning, we use uneven strides based on aspect ratio. The strides $s_h$ and $s_w$ vary, with a smaller stride in the shorter dimension. We use a fixed patch size $p$ of 16. The total patch number $n$ is calculated as $\left(\frac{H - p}{s_h} + 1\right) \times \left(\frac{W - p}{s_w} + 1\right)$.
\begin{figure}
    \centering
\includegraphics[width=0.48\textwidth]{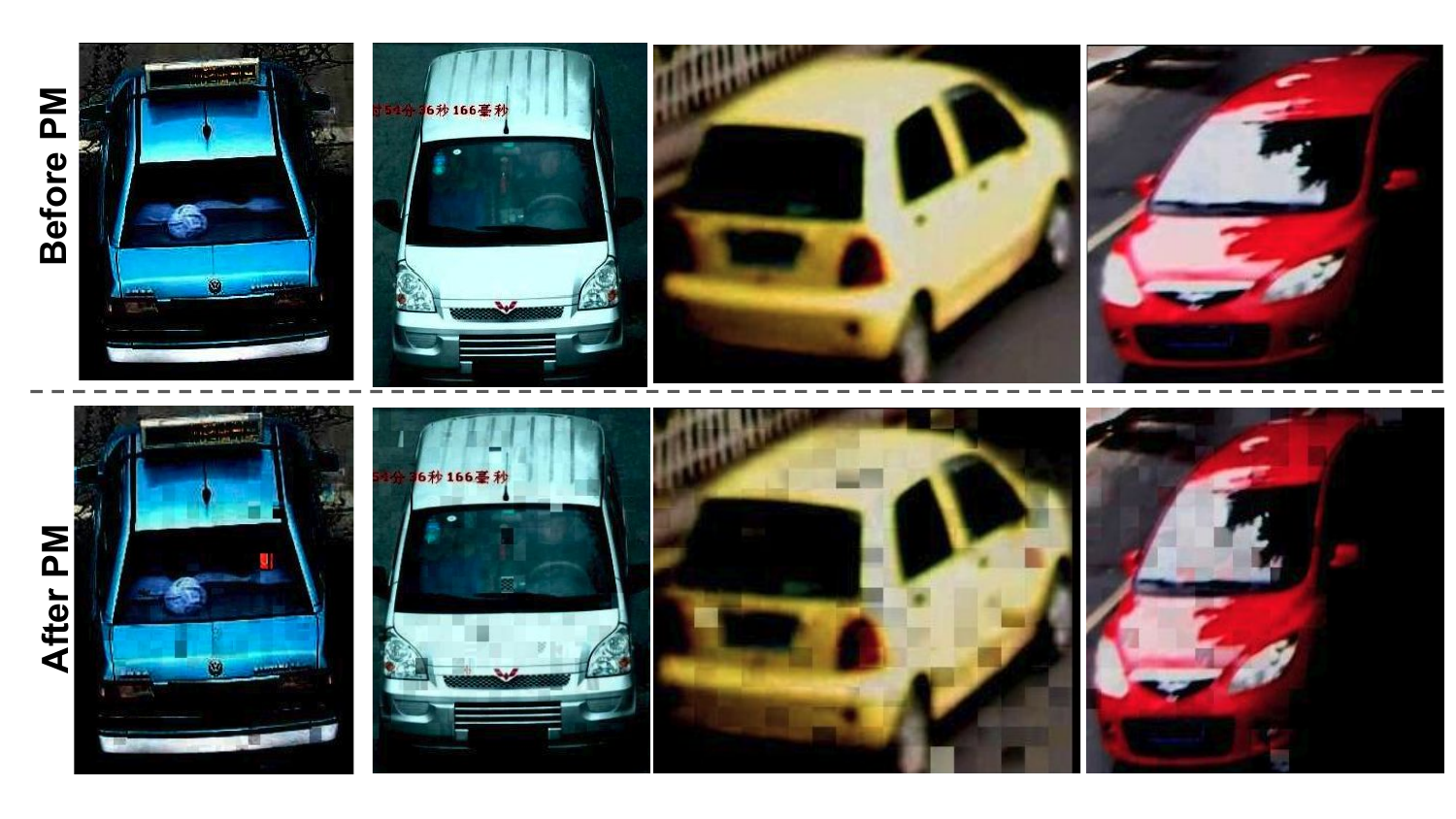} 
    \vspace{-2mm}
    \caption{\textit{Examples from the VehicleID (first two columns) and VeRi-776 (last two columns) test datasets show the impact of intra-image patch mixup (PM). This method blends image parts based on attention-driven distances, enhancing complexity to boost model robustness and reduce overfitting. The top row shows images without PM, while the bottom row includes images processed with PM.
    }}
    \label{fig:aug}
    \vspace{2mm}
\end{figure}

\smallskip
\noindent
\textbf{Patch Mixup Intra-Image Module.} We propose an intra-image data augmentation method where each patch has a probability to mix with another randomly chosen patch from the same image. Mixup weights depend on the spatial distance between patches, with closer patches getting higher weights. Let $A$ denote the original patches with a dimension of $n$ patches, and their positions are $(x_i, y_i)$ for the $i$-th patch. The Euclidean distance between patches $i$ and $j$ is:
\begin{equation}
d(i, j) = \sqrt{(x_i - x_j)^2 + (y_i - y_j)^2}.
\label{d_dis}
\end{equation}

The patch distance matrix \( D \) is:
\begin{equation}
D = \begin{bmatrix}
d(1, 1) & d(1, 2) & \cdots & d(1, n) \\
d(2, 1) & d(2, 2) & \cdots & d(2, n) \\
\vdots & \vdots & \ddots & \vdots \\
d(n, 1) & d(n, 2) & \cdots & d(n, n)
\end{bmatrix}.
\label{d_dis_matrix}
\end{equation}

Given \(D\), the attention score of matrix \(A\) is:
\begin{equation}
A = \frac{1}{1 + D*p}
\label{attention_score}
\end{equation}

Here, $A_{ij}$ increases as $D_{ij}$ decreases, indicating that closer patches have more influence. After shuffling patches, the new mixup weights matrix $A'$ is:
\begin{equation}
A' = A[I_c, I'_{c'}]
\label{attention_score-adjusted}
\end{equation}

Here, $c$ is the index of a patch from $I$, and $c'$ is the index from shuffled patches $I'$. Assuming the original patches $A$ are non-overlapping, and $D$ is calculated with $D = 16 \times D$, the final patches $O$ are given by:
\begin{equation}
O = (1 - A') \times I + A' \times I' 
\end{equation}

Several image samples before and after Patch Mixing (PM) data augmentation are shown in Fig. \ref{fig:aug}. The ablation study results are presented in Table \ref{tab:basic-res}. The pseudocode of this method is shown in Algorithm \ref{alg:patch_mix}.

\begin{algorithm}[t]
\caption{ViT with Intra-Image Patch Mixup}
\label{alg:patch_mix}
\begin{algorithmic}[1]
\State \textbf{Input:} $\mathbf{I}$ (input image batch), $s$ (stride), $ARR$ (aspect ratio range), $p_I$ (image mixup percentage), $p_P$ (patch shuffle percentage), $A$ (original attention score function), $A'$ (new attention score function after shuffle)
\State \textbf{Output:} $\mathbf{O}$ (output image batch)
\State $\mathbf{O} \gets \mathbf{I}$  \Comment{Initialize output with input images}
\State $P \gets (p_h, p_w)$ \Comment{Patch size}
\State $n_x \gets \frac{W - p_w}{s_w} + 1$, $n_y \gets \frac{H - p_h}{s_h} + 1$

\State $V \gets$ indices of images in $\mathbf{I}$ where aspect ratio $AR \in ARR$

\If{$V \neq \emptyset$} 
    \State $M \gets$ select random $p_I\%$ of indices from $V$ \Comment{Select $p_I\%$ of eligible images for patch mixup}
    \For{each $b \in M$}
        \State $P_{\text{shuffle}} \gets$ select random $p_P\%$ of patches from image $b$
        
        \State $A_b \gets A(\mathbf{I}[b])$ \Comment{Generate original attention score for image $b$ using Eq. (\ref{attention_score}})
        
        \State $\mathbf{I'}[b] \gets \text{shuffle}(\mathbf{I}[b])$ \Comment{Shuffle selected patches in image $b$}
        
        \State $A'_b \gets A'(\mathbf{I'}[b])$ \Comment{Generate new attention score after shuffle using Eq. (\ref{attention_score-adjusted}})

        \For{each $(i,j) \in n_y \times n_x$}
            \State $c \gets i \times n_x + j$
            \If{$\mathbf{I}[b][c] \in \mathbf{I'}[b] $}
                \State $\mathbf{O}[b][c] \gets (1-A'_b[c']) \mathbf{I}[b][c] + A'_b[c'] \mathbf{I'}[b][c']$ \Comment{Apply patch mixup with original and new attention scores}
            \EndIf
        \EndFor
    \EndFor
\EndIf

\State \Return $\mathbf{O}$
\end{algorithmic}
\end{algorithm}

\section{Experiments}
\subsection{Experiment Settings}
\smallskip
\noindent
\textbf{Datasets and Evaluation Metrics.} Two popular vehicle ReID datasets, VeRi-776 \cite{zheng2020vehiclenet} and VehicleID \cite{liu2016deep}, are used. We evaluate performance using mAP and CMC metrics, focusing on Rank-1 (R1), Rank-5 (R5), and Rank-10 (R10) accuracy.

\smallskip
\noindent
\textbf{Implementation Details.} We standardized image heights to 224 pixels for VeRi-776 and 384 pixels for VehicleID, using $16 \times 16$ patches with strides of 16 and 12. Aspect ratios were set to $[1.0, 0.95, 1.33]$ for VeRi-776 and $[1.0, 0.80, 1.03]$ for VehicleID. Testing included the full VeRi-776 set and the largest VehicleID subset. Preprocessing involved flipping, padding, cropping, and erasing. For VeRi-776, 75\% of images with aspect ratios within the defined range underwent intra-image patch-mixup, with 25\% mixup at the pixel level. For VehicleID, PM augmentation was applied to 75\% of patches. Experiments were run on four NVIDIA RTX A6000 GPUs with FP16 training, using SGD with 0.1 momentum, 1e-4 weight decay, and a batch size of 128 over 120 epochs. The PM module was used only during training.

\subsection{Results}
\smallskip
\noindent
\textbf{Major Results and Comparison with State-of-the-Art Methods.} Our experimental results (Table \ref{tab:basic-res}) show that on VeRi-776, a non-square input of $224 \times 298$ improves mAP by 4.6\%, with feature fusion enhancing it by 6.5\%. On VehicleID, a $384 \times 396$ input increases mAP by 0.6\%, and feature fusion further boosts mAP by 0.8\% and R1 by 1.3\%. Compared to recent ViT-based methods (Table \ref{tab:2}), our model surpasses pure ViT by 2.5\% mAP on VeRi-776. On the largest VehicleID test set, it achieves 91.0\% mAP, 86.3\% R1, and 97.4\% R5, outperforming pure ViT by 8.4\% in R1. The additional inference time is minimal: 0.005 seconds for VehicleID and 0.002 seconds for VeRi-776.

\begin{figure}[h!]
    \centering
    \begin{minipage}[b]{0.48\textwidth}
        \centering
        \includegraphics[width=0.9\textwidth]{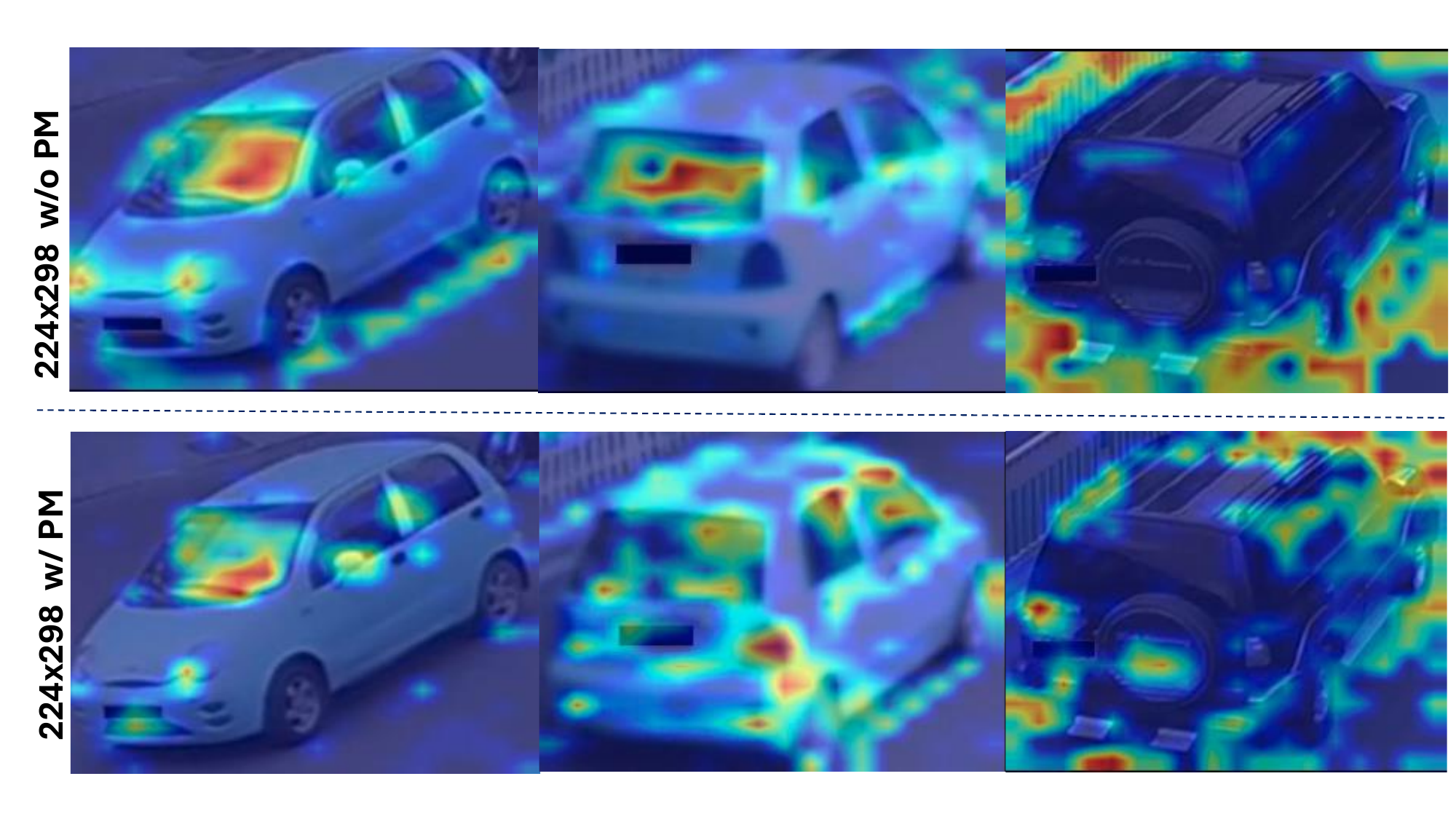}
        \caption{\textit{VeRi-776 attention maps without (top) and with PM module (bottom).}}
        \label{fig:atten-veri}
    \end{minipage}
    \hspace{0.02\textwidth}
    \begin{minipage}[b]{0.48\textwidth}
        \centering
        \includegraphics[width=1.0\textwidth]{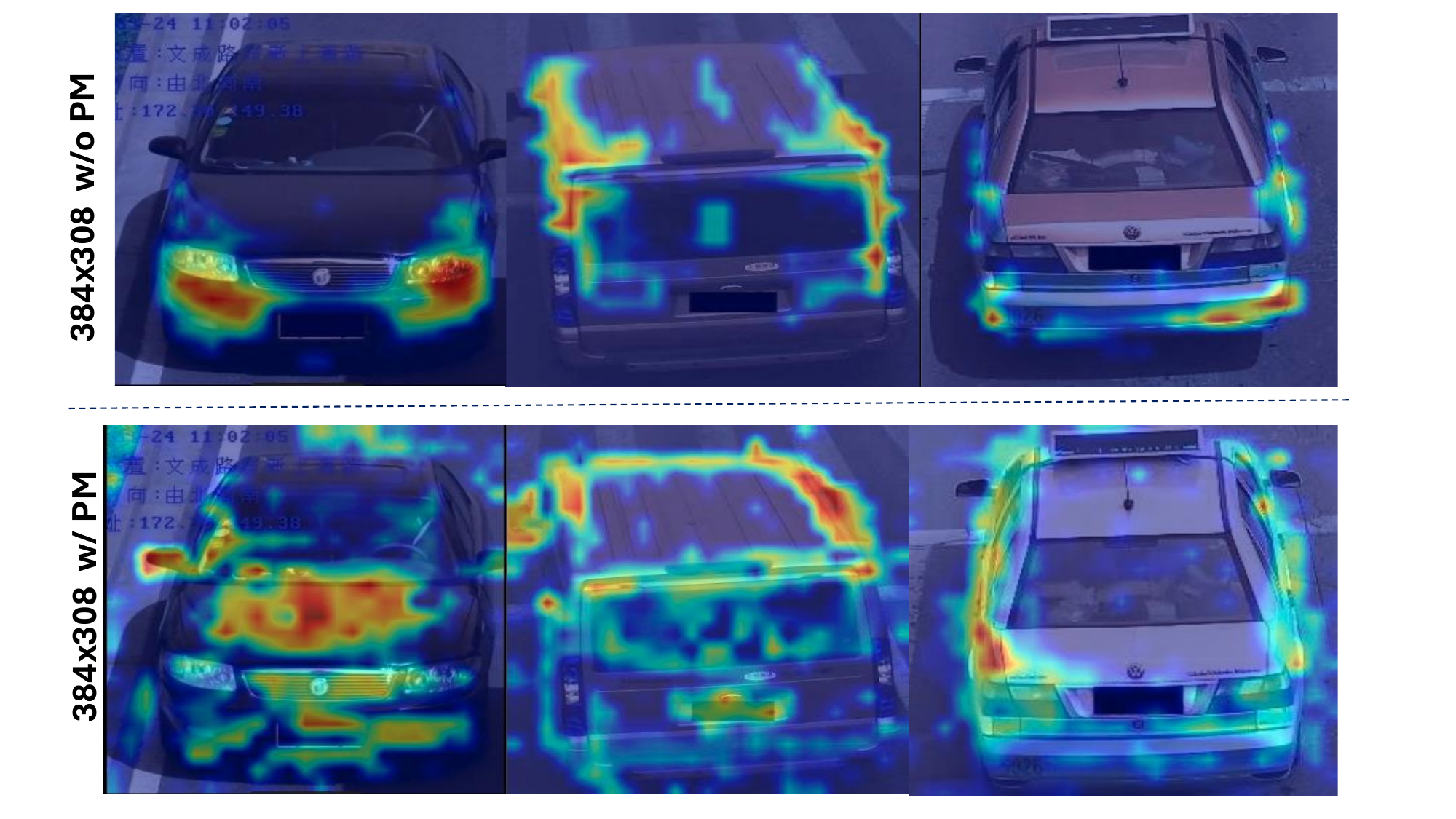}
        \caption{\textit{VehicleID attention maps without (top) and with PM module (bottom).}}
        \label{fig:atten-veh}
    \end{minipage}
\end{figure}

To explore the impact of the PM module, we use Grad-CAM visualizations (Figs. \ref{fig:atten-veri} and \ref{fig:atten-veh}) to highlight model attention. The Intra-image Patch Mixup improves focus on relevant objects, particularly for VeRi-776, and increases robustness to variations in view and aspect ratios. By leveraging self-attention, our method enhances spatial relationships and strengthens the learning of global features from the class token in ViTs.


\begin{table}[ht]
\centering
\caption{The basic results of our method. (w/ and w/o mean `with' and `without')}
\resizebox{\columnwidth}{!}{%
\begin{tabular}{|l|c|c|c|c|c|}
\hline
\textbf{Dataset} & \textbf{Image Size} & \textbf{Aspect Ratio} & \textbf{PM} & \textbf{mAP(\%)} & \textbf{R1(\%)} \\ \hline
\multirow{5}{*}{VeRi776} & 224x224 & 1.0 & w/o & 74.9 & 95.0 \\ \cline{2-6} 
 & 224x224 & 1.0 & w/ & 77.8 & 96.1 \\ \cline{2-6} 
 & 224x212 & 0.95 & w/o & 75.5 & 95.2 \\ \cline{2-6} 
 & 224x212 & 0.95 & w/ & 78.9 & 96.1 \\ \cline{2-6} 
 & 224x298 & 1.33 & w/o & 78.5 & 96.5 \\ \cline{2-6} 
 & 224x298 & 1.33 & w/ & 79.5 (\textbf{4.6}$\uparrow$) & 97.0 \\ \cline{2-6} 
 & \multicolumn{2}{|c|}{\textbf{Fusing}} & w/o & 78.1 & 96.1 \\ \cline{4-6} 
 & \multicolumn{2}{|c|}{} & w/ & 81.4 (\textbf{6.5$\uparrow$}) & 97.0 (\textbf{2.0$\uparrow$}) \\ \hline
\multirow{5}{*}{VehicleID} & 384x384 & 1.0 & w/o & 90.2 & 85.0 \\ \cline{2-6} 
 & 384x384 & 1.0 & w/ & 90.8 (0.6$\uparrow$) & 85.7 \\ \cline{2-6} 
 & 384x308 & 0.80 & w/o & 89.9 & 84.1 \\ \cline{2-6} 
 & 384x308 & 0.80 & w/ & 89.6 & 83.8 \\ \cline{2-6} 
 & 384x396 & 1.03 & w/o & 87.8 & 81.7 \\ \cline{2-6} 
 & 384x396 & 1.03 & w/ & 90.2 & 85.2 \\ \cline{2-6} 
 & \multicolumn{2}{|c|}{\textbf{Fusing}} & w/o & 90.1 & 84.8 \\ \cline{4-6} 
 & \multicolumn{2}{|c|}{} & w/ & 91.0 (\textbf{0.8}$\uparrow$) & 86.3 (\textbf{1.3}$\uparrow$) \\ \hline
\end{tabular}%
}
\label{tab:basic-res}
\end{table}

\begin{table}[ht]
\centering
\caption{Comparison of state-of-the-art \textbf{transformer-based} approaches in vehicle ReID on the datesets VeRi-776 and VehicleID (largest test set).}
\resizebox{\columnwidth}{!}{%
\begin{tabular}{|l|c|c|c|c|c|}
\hline
\multirow{2}{*}{\textbf{Model}} & \multicolumn{3}{c|}{\textbf{VeRi-776}} & \multicolumn{2}{c|}{\textbf{VehicleID (Large)}} \\ \cline{2-6}
 & \textbf{mAP} & \textbf{R1(\%)} & \textbf{R5(\%)} & \textbf{R1(\%)} & \textbf{R5(\%)} \\ \hline
ViT \cite{dosovitskiy2010image}& 78.9 & 96.8 & 98.4 & 77.9 & 92.4 \\ \hline
GiT \cite{shen2023git}& 80.3 & 96.9 & - & 77.9 & - \\ \hline
Vit-ReID \cite{du2023vit} & 80.7 & 96.6 & - & 80.5 & 93.4 \\ \hline
SOFCT \cite{yu2023semantic}& 80.0 & 96.8 & 98.8 & 77.8 & - \\ \hline
SFMNet \cite{li2023sfmnet} & 80.0 & 96.8 & 98.1 & 78.7 & 77.7 \\ \hline
\textbf{Ours} & \textbf{81.4} & \textbf{97.0} & 98.5 & \textbf{86.3} & \textbf{97.4} \\ \hline
\end{tabular}%
}
\label{tab:2}
\end{table}

\section{Conclusion }
In this paper, we demonstrate that fusing ViT-based vehicle ReID models trained on varied aspect ratios significantly enhances robustness and performance. Uneven stride patching preserves spatial structure, while our intra-image Patch Mixup with random sampling improves generalization. These promising results offer new insights into the relationship between aspect ratio, self-attention in vision transformers, and feature generation in ReID tasks. Although this fusion approach outperforms baselines on VeRi-776 and VehicleID datasets, it may increase inference time, which could be mitigated through network pruning. Future work will focus on developing efficient ReID models for diverse aspect ratios and comparing intra-image Patch Mixup with other data augmentation methods.
\vfill\pagebreak

\bibliography{strings.bib}
\bibliographystyle{IEEEtran}

\end{document}